%% file: _main.tex
\input{_constants}

\documentclass[10pt,twocolumn,letterpaper]{article}
\input{cvpr_header}

\usepackage[accsupp]{axessibility}

\begin{document}
\title{\paperTitle}
\author{\authorBlock}
\maketitle

\input{00_abstract}
\input{01_intro}
\input{02_related}
\input{03_method}

\input{04_experiment}
\input{10_conclusion}

{\small
\bibliographystyle{ieee_fullname}
\bibliography{11_references}
}

\clearpage
\section*{Appendix}
\input{12_appendix}
\end{document}

%% file: _constants.tex
\def\cvprPaperID{3843}
\def\confName{CVPR}
\def\confYear{2023}

\def\paperTitle{gSDF: Geometry-Driven Signed Distance Functions \\ for 3D Hand-Object Reconstruction}

\def\authorBlock{
    Zerui Chen \qquad
    Shizhe Chen \qquad
    Cordelia Schmid \qquad
    Ivan Laptev \\
    Inria, \'Ecole normale sup\'erieure, CNRS, PSL Research Univ., 75005 Paris, France\\
    {\tt\small firstname.lastname@inria.fr}\\
    {\tt\small \url{https://zerchen.github.io/projects/gsdf.html}}
}


%% file: cvpr_header.tex
\usepackage[pagenumbers]{cvpr}

\usepackage{graphicx}
\usepackage{amsmath}
\usepackage{amssymb}
\usepackage{booktabs}
\usepackage{color}
\usepackage{soul}

\input{_macros}  

\usepackage{xr-hyper}

\makeatletter
\newcommand*{\addFileDependency}[1]{
  \typeout{(#1)}
  \@addtofilelist{#1}
  \IfFileExists{#1}{}{\typeout{No file #1.}}
}

\makeatother

\usepackage[pagebackref,breaklinks,colorlinks]{hyperref}
\usepackage[capitalize]{cleveref}
\crefname{section}{Sec.}{Secs.}
\crefname{table}{Table}{Tables}
\crefname{figure}{Fig.}{Figs.}

%% file: _macros.tex
\usepackage{times}
\usepackage{microtype}
\usepackage{epsfig}
\usepackage{dsfont}
\usepackage{float}
\usepackage{placeins}
\usepackage{color, colortbl}
\usepackage{stfloats}
\usepackage{enumitem}
\usepackage{tabularx}
\usepackage{xstring}
\usepackage{multirow}
\usepackage{xspace}
\usepackage{url}
\usepackage{subcaption}
\usepackage{xcolor}
\usepackage[hang,flushmargin]{footmisc}
\usepackage{multirow}

% ...

\newcommand{\R}[1]{{%
    \textbf{%
        \ifstrequal{#1}{1}{\textcolor{red}{R#1}}{%
        \ifstrequal{#1}{2}{\textcolor{blue}{R#1}}{%
        \ifstrequal{#1}{3}{\textcolor{magenta}{R#1}}{%
        \ifstrequal{#1}{4}{\textcolor{teal}{R#1}}{%
                           \textcolor{cyan}{R#1}%
        }}}}%
    }%
}}

%% file: 00_abstract.tex
\begin{abstract}
Signed distance functions (SDFs) is an attractive framework that has recently shown promising results for 3D shape reconstruction from images. SDFs seamlessly generalize to different shape resolutions and topologies but lack explicit modelling of the underlying 3D geometry. In this work, we exploit the hand structure and use it as guidance for SDF-based shape reconstruction. In particular, we address reconstruction of hands and manipulated objects from monocular RGB images. To this end, we estimate poses of hands and objects and use them to guide 3D reconstruction. More specifically, we predict kinematic chains of pose transformations and align SDFs with highly-articulated hand poses. We improve the visual features of 3D points with geometry alignment and further leverage temporal information to enhance the robustness to occlusion and motion blurs. We conduct extensive experiments on the challenging ObMan and DexYCB benchmarks and demonstrate significant improvements of the proposed method over the state of the art.

\end{abstract}

%% file: 01_intro.tex
\section{Introduction}
\label{sec:intro}
Understanding how hands interact with objects is becoming increasingly important for widespread applications, including virtual reality, robotic manipulation and human-computer interaction. 
Compared to 3D estimation of sparse hand joints~\cite{zimmermann2017learning,tang2014latent,iqbal2018hand,moon2018v2v,tekin2019h+}, joint reconstruction of hands and object meshes~\cite{hasson2019learning,karunratanakul2020grasping,yang2021cpf,hampali2022keypoint,chen2022alignsdf} provides rich information about hand-object interactions and has received increased attention in recent years. 

\begin{figure}[t]
  \centering
  \includegraphics[trim={5cm 8.5cm 7cm 0cm},clip,width=0.5\textwidth]{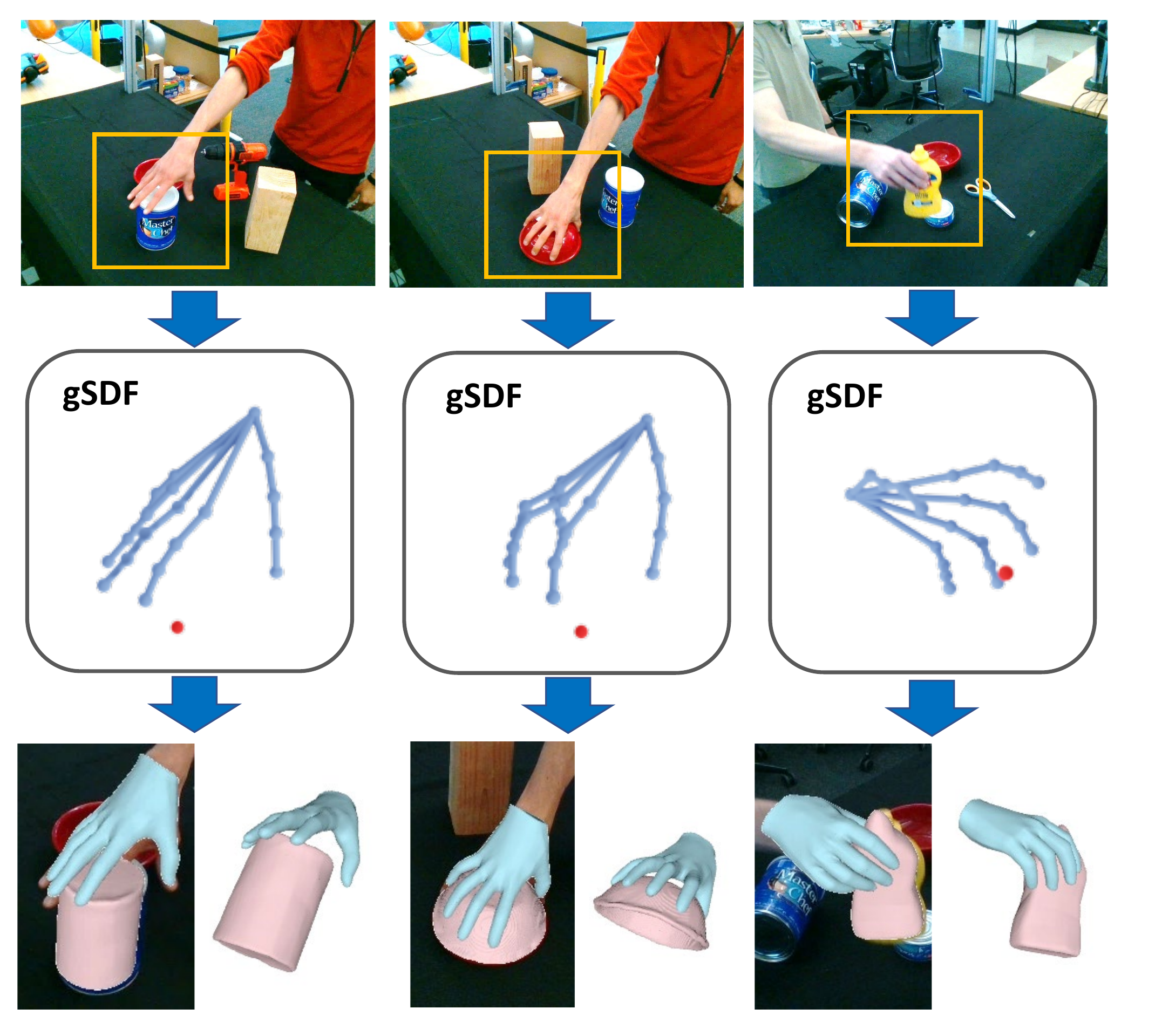}
  \caption{We aim to reconstruct 3D hand and object meshes from monocular images \emph{(top)}. Our method gSDF \emph{(middle)} first predicts 3D hand joints (blue) and object locations (red) from input images. 
  We use estimated hand poses and object locations to incorporate strong geometric priors into SDF by generating hand- and object-aware kinematic features for each SDF query point. 
  Our resulting gSDF model generates accurate results for real images with various objects and grasping hand poses \emph{(bottom)}. 
  }
  \label{teaser}
  \vspace{-0.3cm}
\end{figure}

To reconstruct high-quality meshes, some recent works~\cite{hampali2020honnotate,chao2021dexycb,yang2022oakink} explore multi-view image inputs. Multi-view images, however, are less common both for training and testing scenarios. In this work, we focus on a more practical and user-friendly setting where we aim to reconstruct hand and object meshes from monocular RGB images. Given the ill-posed nature of the task, many existing methods~\cite{hasson2019learning,hasson2020leveraging,tse2022collaborative,yang2021cpf,cao2021reconstructing} employ parametric mesh models (\emph{e.g.}, MANO~\cite{MANO:SIGGRAPHASIA:2017}) to impose prior knowledge and reduce ambiguities in 3D hand reconstruction. MANO hand meshes, however, have relatively limited resolution and can be suboptimal for the precise capture of hand-object interactions.

To reconstruct detailed hand and object meshes, another line of efforts~\cite{karunratanakul2020grasping, chen2022alignsdf} employ signed distance functions (SDFs). Grasping Field~\cite{karunratanakul2020grasping} makes the first attempt to model hand and object surfaces using SDFs. However, it does not explicitly associate 3D geometry with image cues and has no prior knowledge incorporated in SDFs, leading to unrealistic meshes. AlignSDF~\cite{chen2022alignsdf} proposes to align SDFs with respect to global poses (\emph{i.e.}, the hand wrist transformation and the object translation) and produces improved results. However, it is still challenging to capture geometric details for more complex hand motions and manipulations of diverse objects, which involve the articulation of multiple fingers.

To address limitations of prior works, we propose a geometry-driven SDF (gSDF) method that encodes strong pose priors and improves reconstruction by disentangling pose and shape estimation
(see Figure~\ref{teaser}). 
To this end, we first predict sparse 3D hand joints from images and derive full kinematic chains of local pose transformations from joint locations using inverse kinematics. 
Instead of only using the global pose as in \cite{chen2022alignsdf}, we optimize SDFs with respect to poses of all the hand joints, which leads to a more fine-grained alignment between the 3D shape and articulated hand poses. In addition, we project 3D points onto the image plane to extract geometry-aligned visual features for signed distance prediction. The visual features are further refined with spatio-temporal contexts using a transformer model to enhance the robustness to occlusions and motion blurs.

We conduct extensive ablation experiments to show the effectiveness of different components in our approach. The proposed gSDF model greatly advances state-of-the-art accuracy on the challenging ObMan and DexYCB benchmarks. Our contributions can be summarized in three-fold: \emph{(i)} To embed strong pose priors into SDFs, we propose to align the SDF shape with its underlying kinematic chains of pose transformations, which reduces ambiguities in 3D reconstruction. \emph{(ii)} To further reduce the misalignment induced by inaccurate pose estimations, we propose to extract geometry-aligned local visual features and enhance the robustness with spatio-temporal contexts. \emph{(iii)} We conduct comprehensive experiments to show that our approach outperforms state-of-the-art results by a significant margin. 

%% file: 02_related.tex
\section{Related Work}
\label{sec:related}
This paper focuses on jointly reconstructing hands and hand-held objects from RGB images. In this section, we first review previous works on the 3D hand pose and shape estimation. We then discuss relevant works on the joint reconstruction of hands and objects.

{\bf 3D hand pose and shape estimation.} The topic of 3D hand pose estimation has received widespread attention since the 90s~\cite{rehg1994visual,heap1996towards} and has seen significant progress in recent years~\cite{lepetit2020recent_pose_advances,Yuan_2018_CVPR}.
Methods which take RGB images as input~\cite{zimmermann2017learning,iqbal2018hand,tang2014latent,tekin2019h+,moon2018v2v,spurr2021self,xiong2019a2j,sun2018integral,mueller2018ganerated,meng20223d} often estimate sparse 3D hand joint locations from visual data using well-designed deep neural networks. Though these methods can achieve high estimation accuracy, their outputs of 3D sparse joints provide limited information about the 3D hand surface, which is critical in AR/VR applications.
Following the introduction of the anthropomorphic parametric hand mesh model MANO~\cite{MANO:SIGGRAPHASIA:2017}, 
several works~\cite{lv2021handtailor,wang2020rgb2hands,mueller2019real,kulon2019rec,Kulon2020weaklysupervisedmh,baek2019pushing,chen2021camera,boukhayma20193d,hampali2022keypoint,li2022interacting} estimate the MANO hand shape and pose parameters to recover the full hand surface.
However, MANO has a limited mesh resolution and cannot produce fine surface details.
Neural implicit functions~\cite{karunratanakul2021skeleton,corona2022lisa} have the potential to reconstruct more realistic high resolution hand surfaces~\cite{park2019deepsdf,mescheder2019occupancy,chen2019learning}.
In this work, we combine the advantages of sparse, parametric and implicit modelling.
We predict sparse 3D joints accurately from images and estimate the MANO parameters using inverse kinematics. We then optimize neural implicit functions with respect to underlying kinematic structures and reconstruct realistic meshes.

{\bf 3D hand and object reconstruction.} Joint reconstruction of hand and object meshes provides a more comprehensive view about how hands interact with manipulated objects in the 3D space and has received more attention in the past few years. Previous works often rely on multi-view correspondence~\cite{chao2021dexycb,yang2022oakink,hampali2020honnotate,ballan2012motion,oikonomidis2011full,wang2013video} or additional depth information~\cite{hamer2009tracking,hamer2010object,sridhar2016real,tsoli2018joint,tzionas20153d} to approach this task.
In this work, we focus on a more challenging setting and perform a joint reconstruction from monocular RGB images.
Given the ill-posed nature of this problem, many works~\cite{hasson2021towards,hasson2020leveraging,yang2021cpf,tse2022collaborative,cao2021reconstructing,hasson2019learning,yang2022artiboost,hampali2022keypoint} deploy MANO, which encodes hand prior knowledge learned from hand scans, to reconstruct hand meshes. 
To further simplify the object reconstruction task, several works~\cite{yang2022artiboost,yang2021cpf,hampali2022keypoint} make a strong assumption that the ground-truth object model is available at test time.
Our work and some previous efforts~\cite{hasson2019learning,karunratanakul2020grasping,chen2022alignsdf} relax this assumption and assume unknown object models.
Hasson \emph{et al.}~\cite{hasson2019learning} employ a differentiable MANO layer to estimate the hand shape and AtlasNet~\cite{groueix2018papier} to reconstruct the manipulated object. However, both MANO and AtlasNet can only produce meshes of limited resolution, which prevents the modelling of detailed contacts between hands and objects.
To generate more detailed surfaces, Karunratanakul~\emph{et al.}~\cite{karunratanakul2020grasping} introduce grasping fields and propose to use SDFs to reconstruct both hand and object meshes. However, such a model-free approach does not capture any prior knowledge about hands or objects, which can lead to predicting unrealistic 3D geometry. To mitigate this, Ye \emph{et al.}~\cite{ye2022s} propose to use hand poses estimated from an off-the-shelf model to help reconstruct the hand-held object mesh. The main difference with our work is that we jointly reconstruct hand meshes and object meshes using our proposed model, which is more challenging. 
Also, in addition to using hand poses to help capture the object shapes, we predict object poses and show their benefits for SDF-based object reconstruction. Another work AlignSDF~\cite{chen2022alignsdf} optimizes SDFs with respect to estimated hand-object global poses and encodes pose priors into SDFs. In addition to using global poses as a guide for SDFs, we propose to learn SDFs from the full kinematic chains of local pose transformations, and achieve a more precise alignment between the 3D shape and the underlying poses.
To further handle hard cases  induced by occlusion or motion blur where pose estimations are inaccurate, we leverage a transformer to accumulate corresponding image features from multiple frames and benefit the geometry recovery.

%% file: 03_method.tex
\section{Method}
\label{sec:method}
This section presents our geometry-driven SDF (gSDF) method for 3D hand and object reconstruction from monocular RGB images.
We aim to learn two signed distance functions $\text{SDF}_{hand}$ and $\text{SDF}_{obj}$ to implicitly represent 3D shapes for the hand and the object.
The $\text{SDF}_{hand}$ and $\text{SDF}_{obj}$ map a query 3D point $x \in \mathbb{R}^3$ to a signed distance from the hand surface and object surface, respectively.
The Marching Cubes algorithm~\cite{lorensen1987marching} can thus be employed to reconstruct the hand and the object from ${\rm SDF}_{hand}$ and ${\rm SDF}_{obj}$.

\begin{figure*}[t]
  \centering
  \includegraphics[width=1\linewidth]{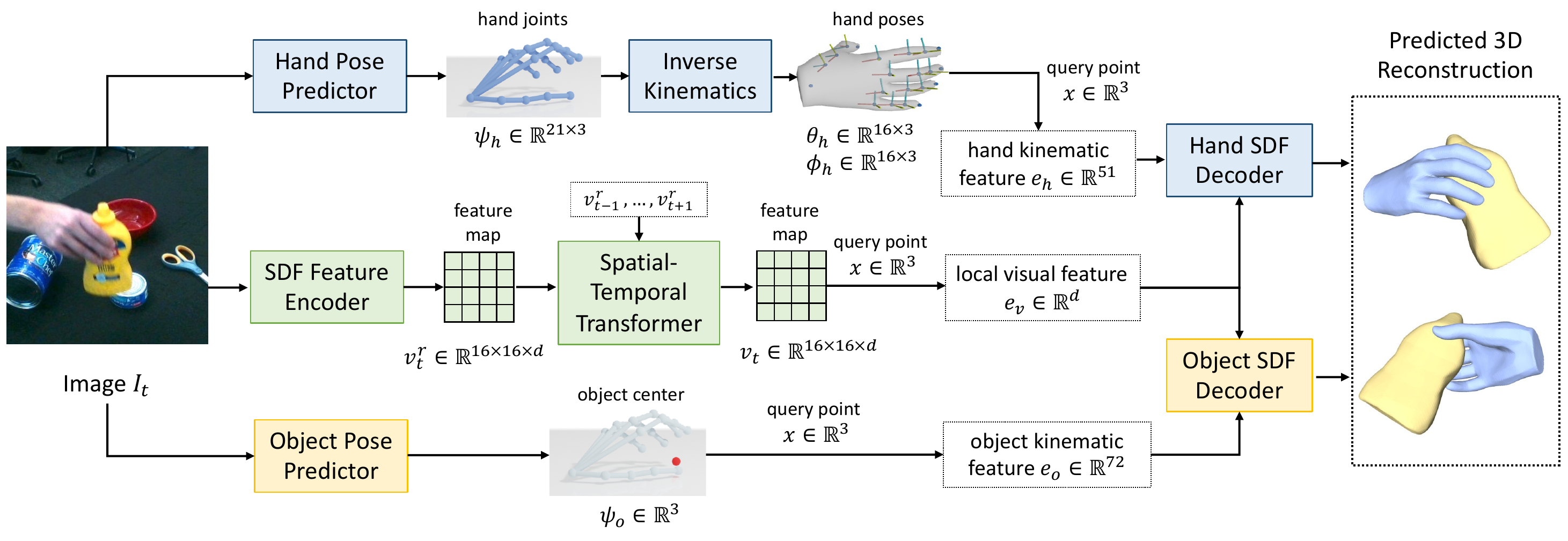}
  \caption{The overview of our proposed single-frame model. Our method reconstructs realistic hand and object meshes from a single RGB image. Marching Cubes algorithm~\cite{lorensen1987marching} is used at test time to extract meshes.
  \vspace{-.1cm}}
  \label{fig:method}
\end{figure*}

\subsection{Overview of gSDF}
\label{sec:method_overview}

Figure~\ref{fig:method} illustrates the overview of our gSDF reconstruction approach.
Given an image $I_t$, we extract two types of features to predict the signed distance for each query point $x$, namely kinematic features and visual features.

The kinematic feature encodes the position of $x$ under the coordinate system of the hand or the object, which can provide strong pose priors to assist SDF learning. Since the feature is based on canonical hand and object poses, it helps to {\em disentangles} shape learning from pose learning.

\begin{figure}[!t]
  \centering
  \includegraphics[trim={0cm 1cm 10.3cm 0cm},clip, width=.47\textwidth]{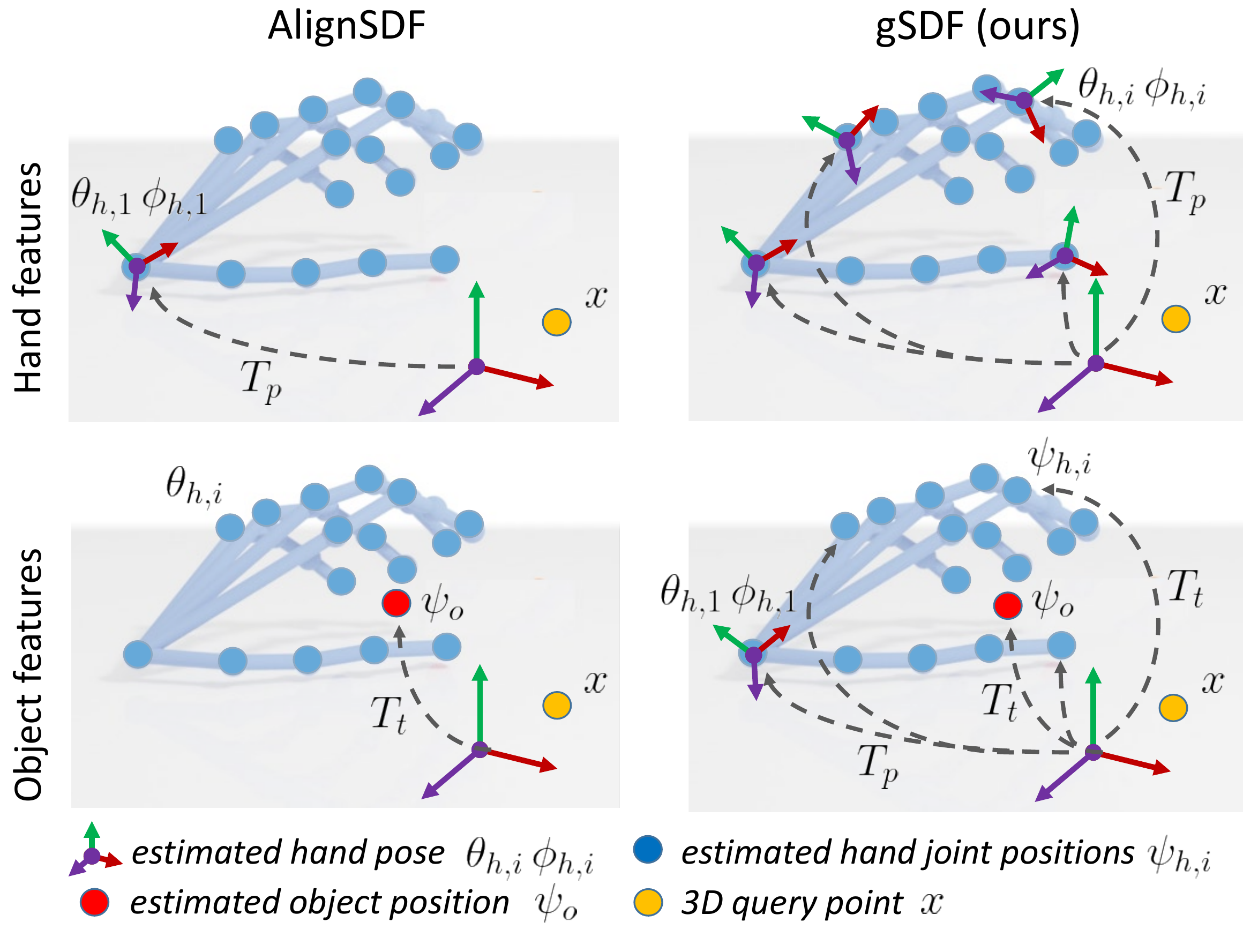}
  \vspace{-0.1cm}
  \caption{We define hand and object features by transforming queries $x$ into hand- and object-centered coordinate systems. Compared to AlignSDF~\cite{chen2022alignsdf} (left), each hand joint in our method defines its own coordinate frame.}
  \label{fig:method_transform}
\end{figure}

The existing work~\cite{ye2022s} proposes to use hand poses for reconstructing object meshes but does not consider using pose priors to reconstruct hand meshes. 
Another work \cite{chen2022alignsdf} only deploys coarse geometry in terms of the hand wrist object locations, which fails to capture fine-grained details. In this work, we aim to strengthen the kinematic feature with geometry transformation of $x$ to poses of all the hand joints (see Figure~\ref{fig:method_transform}) for both the hand and the object reconstruction.
However, it is challenging to directly predict hand pose parameters~\cite{bregier2021deep,zhou2019continuity,kolotouros2019learning}.
To improve the hand pose estimation, we propose to first predict sparse 3D joint locations $j_h$ from the image and then use inverse kinematics to derive pose transformations $\theta_h$ from the predicted joints. In this way, we are able to obtain kinematic features $e_h$ and $e_o$ for the hand and the object respectively.

The visual feature encodes the visual appearance for the point $x$ to provide more shape details.
Prior works \cite{chen2022alignsdf,karunratanakul2020grasping} use the same global visual feature for all the points, \eg, averaging the feature map of a SDF feature encoder on the spatial dimension.
Such global visual features suffers from imprecise geometry alignment between a point and its visual appearance.
To alleviate the limitation, inspired by~\cite{saito2019pifu}, we apply the geometry transformation to extract aligned local visual features. Moreover, to address hard cases with occlusions and motion blur in a single image $I_t$, we propose to enhance the local visual feature with its temporal contexts from videos using a spatio-temporal transformer. We denote the local visual feature of a point as $e_v$. Finally, we concatenate the kinematic feature and local visual feature to predict the signed distance for $x$:
\begin{equation}
\begin{split}
&{\rm SDF}_{hand}(x) = f_{h}([e_{v}; e_{h}]), \\
&{\rm SDF}_{object}(x) = f_{o}([e_{v}; e_{o}]),
\end{split}
\label{sdf}
\end{equation} 
where $f_{h}$ and $f_{o}$ are the hand SDF decoder and the object SDF decoder respectively. 

In the following, we first present the proposed geometry-driven kinematic feature and visual feature encodings in Section~\ref{sec:method_kinematic_embed} and~\ref{sec:method_visual_embed} respectively. 
Then, in Section~\ref{sec:method_backbone} we introduce different strategies of sharing image backbones for hand and object pose predictors as well as the SDF feature encoder. 
Finally, the training strategy of our model is described in Section~\ref{sec:method_train}.

\subsection{Kinematic Feature Encoding}

\label{sec:method_kinematic_embed}
\noindent\textbf{Hand and object pose estimation.}
Directly regressing hand pose parameters of MANO from image features \cite{hasson2019learning,hasson2020leveraging,chen2022alignsdf} has proved to be difficult~\cite{bregier2021deep,zhou2019continuity,kolotouros2019learning}. 
In contrast, predicting sparse 3D joint locations is easier and can achieve higher accuracy.
Therefore, we first train a 3D hand joint prediction model which produces volumetric heatmaps~\cite{pavlakos2017coarse,moon2018v2v} for 21 hand joints.
We use a differentiable soft-argmax operator~\cite{sun2018integral} to extract 3D coordinates $\psi_{h} \in \mathbb{R}^{21 \times 3}$ of hand joints from the heatmaps.
We then obtain an analytic solution for hand poses $\theta_{h} \in \mathbb{R}^{16 \times 3}, \phi_{h} \in \mathbb{R}^{16 \times 3}$ from estimated 3D joints $\psi_{h}$ using inverse kinematics, where each $\theta_{h,i} \in \mathbb{R}^3$ and $\theta_{h,i} \in \mathbb{R}^3$ denote the relative pose of $i_{th}$ joint in terms of rotation and translation with respect to its ancestor joint.
Here, we only calculate the rotation and use the default limb lengths provided by the MANO model.
%
Specifically, we first compute the pose of the hand wrist using the template pose defined in MANO, and then follow the hand kinematic chain to solve the pose of other finger joints recursively.
More details are presented in the appendix.

For the object pose estimation, it is often difficult to accurately estimate the rotation of the object since many objects have a high degree of symmetry and are often occluded by hands. We therefore follow~\cite{chen2022alignsdf} and only estimate the center position 
of the object $\psi_{o} \in \mathbb{R}^{3}$ relative to the hand wrist. 

\noindent\textbf{Hand kinematic feature.}
Given the 3D point $x$, we generate the hand kinematic feature $e_{h} \in \mathbb{R}^{51}$ by transforming $x$ into canonical coordinate frames defined by hand joints. 
Figure~\ref{fig:method_transform}(top,right) illustrates the proposed geometry transformation for the hand.
For the $i_{th}$ hand joint pose $\theta_{h,i}, \phi_{h,i}$, the pose transformation $T_{p}(x, \theta_{h,i}, \phi_{h,i})$ to obtain the local hand kinematic feature $e_{h, i} \in \mathbb{R}^{3}$ is defined as
\begin{equation}
\centering
\begin{split}
&G_{h, i} = \prod_{j \in A(i)}\left[\begin{array}{c|c}
\exp \left (\theta_{h, j}\right) & \phi_{h, j} \\
\hline 0 & 1
\end{array}\right], \\
&e_{h, i} = T_{p} (x, \theta_{h,i}, \phi_{h,i}) = \widetilde{H}(G_{h, i}^{-1} \cdot H(x)),
\end{split}
\label{hand_local_embed}
\end{equation}
where $A(i)$ denotes the ordered set of ancestors of the $i_{th}$ joint.
We use \emph{Rodrigues formula} ${\rm exp(\cdot)}$ to convert $\theta_{h, i}$ into the form of a rotation matrix. By traversing the hand kinematic chain, we obtain the global transformation $G_{h, i} \in \mathbb{R}^{4\times 4}$ for the $i_{th}$ joint. Then, we take the inverse of $G_{h, i}$ to transform $x$ into the $i_{th}$ hand joint canonical coordinates. $H(\cdot)$ transforms $x$ into homogeneous coordinates while $\widetilde{H}(\cdot)$ transforms homogeneous coordinates back to Euclidean coordinates. 
Given local kinematic features $e_{h,i}$, the hand kinematic feature $e_{h} \in \mathbb{R}^{51}$ is defined as:
\begin{equation}
\begin{split}
e_{h} = [x, e_{h, 1}, \cdots, e_{h, 16}].
\end{split}
\label{hand_embed}
\end{equation}

\noindent\textbf{Object kinematic feature.}
To obtain geometry-aware SDF for object reconstruction, we propose object kinematic feature $e_{o} \in \mathbb{R}^{72}$. Following~\cite{chen2022alignsdf}, we use estimated object center $\psi_{o}$ to transform $x$ into the object canonical coordinate frame by the translation transformation $x_{oc} = T_t(x,\psi_o)= x - \psi_{o}$. 
As the grasping hand pose also gives hints about the shape of the manipulated object, similar to~\cite{ye2022s} we incorporate the knowledge of hand poses into object reconstruction. 
To this end, for each joint $i$ and its estimated 3D location $\psi_{h,i}$, we transform $x$ by translation as
\begin{equation}
e_{o,i}=T_t(x,\psi_{h,i})= x - \psi_{j,i}.
\end{equation}
Given the importance of the wrist motion for object grasping, 
we also transform $x$ into the canonical coordinate system of the hand wrist $x_{ow} = T_p(x, \theta_{h,1}, \phi_{h,1}) = \widetilde{H}(G_{h, 1}^{-1} \cdot H(x))$, which normalizes the orientation of the grasping and further simplifies the task for the SDF object decoder. 
The object kinematic feature is then defined by $e_{o} \in \mathbb{R}^{72}$ as
\begin{equation}
\begin{split}
e_{o} = [x, x_{oc}, e_{o,1}, \cdots, e_{o,21}, x_{ow}].
\end{split}
\label{object_embed}
\end{equation}
Figure~\ref{fig:method_transform}(bottom,right) illustrates the proposed geometry transformation for the object kinematic feature.

\subsection{Visual Feature Encoding}
\label{sec:method_visual_embed}

\noindent\textbf{Geometry-aligned  visual feature.}
Previous works~\cite{karunratanakul2020grasping,chen2022alignsdf} typically predict signed distances from global image features that lack spatial resolution. 
Motivated by~\cite{saito2019pifu}, we aim to generate geometry-aligned local image features for each input point $x$. 
Assume $v^r_t \in \mathbb{R}^{16 \times 16 \times d}$ is the feature map generated from the SDF feature encoder, \eg a ResNet model~\cite{he2016deep}, where $16\times 16$ is the spatial feature resolution and $d$ is the feature dimension.
We project the 3D input point $x$ to $\hat x$ on the image plane with the camera projection matrix and use bilinear sampling to obtain a local feature $e_v$ from the location on the feature map corresponding to $\hat x$.

\smallskip
\noindent\textbf{Temporaly-enhanced visual feature.}
To improve the robustness of visual features in a single frame $I_t$ from occlusion or motion blur, we propose to exploit temporal information from videos to refine $v^r_t$. Note that due to non-rigid hand motions, we do not assume video frames to contain different views of the same rigid scene.
We make use of the spatial-temporal transformer architecture~\cite{arnab2021vivit,bertasius2021space} to efficiently propagate image features across frames. 
Assume $v^r_{t-1}, \cdots v^r_{t+1}$ are the feature maps from neighboring frames of $I_t$ in a video.
We flatten all the feature maps as a sequence in the spatial-temporal dimension leading to $3 \times 16 \times 16$ tokens fed into the transformer model.
We reshape the output features of the transformer into a feature map again for $I_t$, denoted as $v_t \in \mathbb{R}^{16 \times 16 \times d}$.
By aggregating spatial and temporal information from multiple frames, $v_t$ becomes more robust to the noise and can potentially produce more stable reconstruction results compared to $v^r_t$. 
Our full gSDF model relies on the feature map $v_t$ to compute the local visual feature $e_v$ for the given input point $x$.

\subsection{Image Backbone Sharing Strategy}
\label{sec:method_backbone}

\begin{figure}
     \centering
     \begin{subfigure}[b]{1\linewidth}
         \centering
         \includegraphics[width=\linewidth]{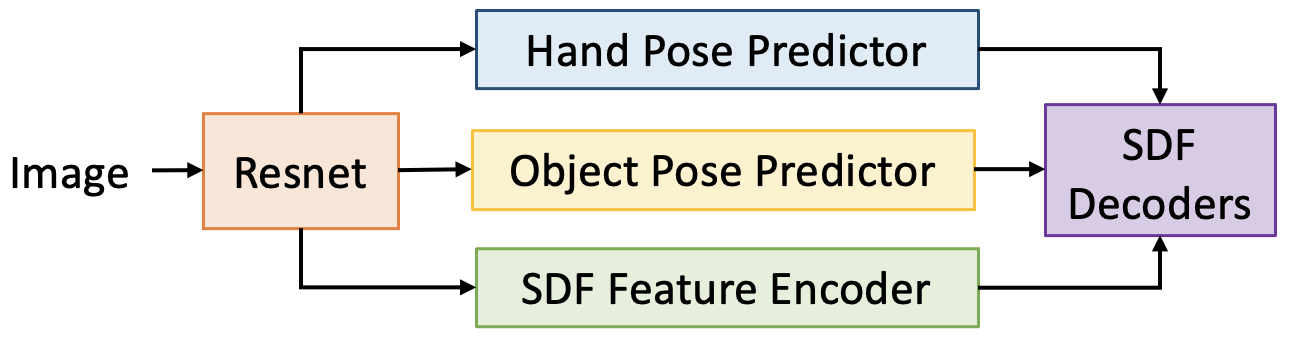}
         \caption{Single backbone.}
         \label{fig:backbone_single}
     \end{subfigure}
     \hfill
     \begin{subfigure}[b]{1\linewidth}
         \centering
         \includegraphics[width=\textwidth]{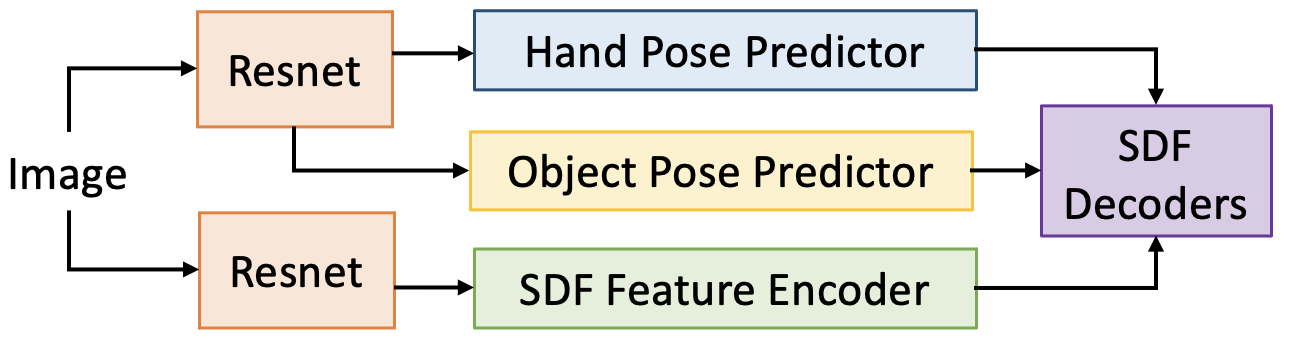}
         \caption{Symmetric backbone.}
         \label{fig:backbone_sym}
     \end{subfigure}
     \hfill
     \begin{subfigure}[b]{1\linewidth}
         \centering
         \includegraphics[width=\textwidth]{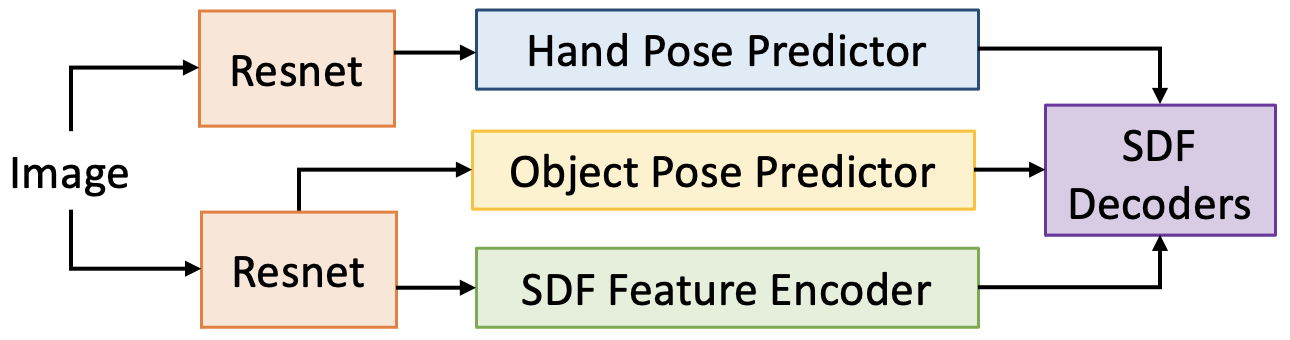}
         \caption{Asymmetric backbone.}
         \label{fig:backbone_asym}
     \end{subfigure}
        \caption{Illustrations of three image backbone sharing strategies.}
     \label{fig:image_backbone_sharing}
\end{figure}

As shown in Figure~\ref{fig:method}, our model contains three branches for hand and object pose estimations as well as for SDF feature encoding.
These different branches may share image backbones which might be beneficial with the multi-task learning.
In this section, we describe three alternative strategies for sharing image backbones in our model.

\noindent\textbf{Single image backbone} (Figure~\ref{fig:image_backbone_sharing}a).
We only employ one single image backbone for both pose and shape predictions.
This is the strategy used in AlignSDF~\cite{chen2022alignsdf}.

\noindent\textbf{Symmetric image backbone} (Figure~\ref{fig:image_backbone_sharing}b).
To disentangle pose and shape learning, we share the image backbone for hand and object pose estimation, but use a different backbone to extract visual features for SDFs learning.

\noindent\textbf{Asymmetric image backbone} (Figure~\ref{fig:image_backbone_sharing}c).
Since hand pose estimation plays a critical role in the task, we use a separate backbone to predict the hand pose, while share the image backbone for object pose predictor and SDF feature encoder.

\subsection{Training}
\label{sec:method_train}

We apply a two-stage training strategy. 
In the first stage, we train the hand pose predictor to predict hand joint coordinates $\psi_{h}$ with $\ell 2$ loss $\mathcal{L}_{hp}$ and an ordinal loss~\cite{pavlakos2018ordinal} $\mathcal{L}_{ord}$ to penalize the case if the predicted depth order between the $i_{th}$ joint and the $j_{th}$ joint is misaligned with the ground-truth relation $\mathds{1}^{ord}_{i,j}$, which are:
\begin{equation}
\mathcal{L}_{hp} = \frac{1}{21} \sum_{i=1}^{21} \Big \| \psi_{h, i} - \hat{\psi}_{h, i} \Big \|^2_2,
\label{eq:hand_pose_loss}
\end{equation}
\begin{equation}
    \mathcal{L}_{ord} =  \sum_{j=2}^{21}\sum_{i=1}^{j-1} \mathds{1}^{ord}_{i,j} \times \Big | (\psi_{h, i} - \psi_{h, j}) \cdot \Vec{n} \Big |,
\label{eq:ord_loss} 
\end{equation} 
where $\Vec{n} \in \mathbb{R}^{3}$ denotes the viewpoint direction. 
We randomly sample twenty virtual views to optimize $\mathcal{L}_{ord}$.
Since the proposed kinematic features are based on the predicted hand joints $\psi_{h}$, we empirically find that pretraining the hand joint predictor in the first stage and then freezing its weights can achieve better performance.

In the second training stage, we learn all the modules except the hand joint predictor in an end-to-end manner.
We use the $\ell 2$ loss $\mathcal{L}_{op}$ to predict the object pose $\psi_o$ as follows:
\begin{equation}
\mathcal{L}_{op} = \Big \| \psi_{o} - \hat{\psi}_{o} \Big \|^2_2
\label{eq:pose_loss}
\end{equation}
where $\hat{\psi}_{o}$ denote the ground-truth location for the object center. 
To train the SDFs, we sample many 3D points around the hand-object surface and calculate their ground-truth signed distances to the hand mesh and the object mesh. 
We use $\ell 1$ loss to optimize the SDF decoders:
\begin{equation}
\begin{split}
&\mathcal{L}_{hsdf} = \Big \| {\rm SDF}_{hand} - \hat{{\rm SDF}}_{hand} \Big \|_1^1, \\
&\mathcal{L}_{osdf} = \Big \| {\rm SDF}_{obj} - \hat{{\rm SDF}}_{obj} \Big \|_1^1, 
\end{split}
\label{loss_sdf}
\end{equation} 
where $\hat{{\rm SDF}}_{hand}$ and $\hat{{\rm SDF}}_{obj}$ denote ground-truth signed distances to the hand and the object, respectively. 
The overall training objective $\mathcal{L}_{shape}$ in the second training stage is:
\begin{equation}
\begin{split}
\mathcal{L}_{shape} = \mathcal{L}_{op} + 0.5\times \mathcal{L}_{hsdf} + 0.5\times \mathcal{L}_{osdf}.
\end{split}
\label{loss_shape}
\end{equation}


%% file: 04_experiment.tex
\section{Experiments}
\label{sec:experiment}
We conduct extensive experiments on two 3D hand-object reconstruction benchmarks to evaluate the effectiveness of our proposed gSDF model.

\subsection{Datasets}
\label{subsec:benchmarks}
\noindent \textbf{ObMan}~\cite{hasson2019learning} is a large-scale synthetic dataset that contains diverse hand grasping poses on a wide range of objects imported from ShapeNet~\cite{chang2015shapenet}. 
We follow previous methods~\cite{park2019deepsdf,karunratanakul2020grasping,chen2022alignsdf,ye2022s} to generate data for SDFs training.
First, we remove meshes that contain too many double-sided triangles, which results in 87,190 hand-object meshes. Then, we fit the hand-object mesh into a unit cube and sample 40,000 points inside the cube. For each sampled point, we compute its signed distance to the ground-truth hand mesh and object mesh, respectively. 
At test time, we report the performance on the whole ObMan test set of 6,285 testing samples.

\noindent \textbf{DexYCB}~\cite{chao2021dexycb} is currently the largest real dataset that captures hand and object interactions in videos. Following~\cite{yang2022artiboost,chen2022alignsdf}, we focus on right-hand samples and use the official s0 split. We follow the same steps as in ObMan to obtain SDF training samples. To reduce the temporal redundancy, we downsample the video data to 6 frames per second, which results in 29,656 training samples and 5,928 testing samples.

\subsection{Evaluation metrics}
\label{subsec:metrics}
We follow prior works to comprehensively evaluate the 3D reconstructions with multiple metrics as below.

\noindent \textbf{Hand Chamfer Distance (${\rm {\bf{CD}}_{\bf{h}}}$)}. We evaluate Chamfer distance (${\rm cm}^{2}$) between our reconstructed hand mesh and the ground-truth hand mesh. We follow previous works~\cite{karunratanakul2020grasping,chen2022alignsdf} to optimize the scale and translation to align the reconstructed mesh with the ground truth and sample 30,000 points on both meshes to compute Chamfer distance. We report the median Chamfer distance on the test set to reflect the quality of our reconstructed hand mesh.

\noindent \textbf{Hand F-score (${\rm {\bf{FS}}_{\bf{h}}}$)}. Since Chamfer distance is vulnerable to outliers~\cite{tatarchenko2019single,ye2022s}, we also report the F-score to evaluate the predicted hand mesh. After aligning the hand mesh with its ground truth, we report F-score at 1mm (${\rm {{FS}}_{{h}}}@1$) and 5mm (${\rm {{FS}}_{{h}}}@5$) thresholds. 

\noindent \textbf{Object Chamfer Distance (${\rm {\bf{CD}}_{\bf{o}}}$)}. Following~\cite{karunratanakul2020grasping,chen2022alignsdf}, we first use the optimized hand scale and translation to transform the reconstructed object mesh. Then, we follow the same process as ${\rm {{CD}}_{{h}}}$ to compute ${\rm {{CD}}_{{o}}}$ (${\rm cm}^{2}$) and evaluate the quality of our reconstructed object mesh.

\noindent \textbf{Object F-score (${\rm {\bf{FS}}_{\bf{o}}}$)}. We follow the previous work~\cite{ye2022s} to evaluate the reconstructed object mesh using F-score at 5 mm (${\rm {{FS}}_{{o}}}@5$) and 10 mm (${\rm {{FS}}_{{o}}}@10$) thresholds.

\noindent \textbf{Hand Joint Error (${\rm {\bf{E}}_{\bf{h}}}$)}. To measure the hand pose estimation accuracy, we compute the mean joint error (${\rm cm}$) relative to the hand wrist over all 21 joints in the form of $\ell 2$ distance.

\noindent \textbf{Object Center Error (${\rm {\bf{E}}_{\bf{o}}}$)}. To evaluate the accuracy of our predicted object translation, we report the $\ell 2$ distance (${\rm cm}$) between the prediction and its ground truth.

Additionally, we report Contact ratio (${\rm {{C}}}_{{r}}$), Penetration depth (${\rm {{P}}}_{{d}}$) and Intersection volume (${\rm {{I}}}_{{v}}$)~\cite{hasson2019learning,yang2021cpf,yang2022artiboost,karunratanakul2020grasping,chen2022alignsdf} to present more details about the interaction between the hand mesh and the object mesh. Please see the appendix for more details.

\subsection{Implementation details}
\label{subsec:details}
\noindent \textbf{Model architecture.} 
We use ResNet-18~\cite{he2016deep} as our image backbone.
For hand and object pose estimation, we adopt volumetric heatmaps of spatial resolution $64 \times 64 \times 64$ to localize hand joints and the object center in 3D space. 
For the spatial-temporal transformer, we use 16 transformer layers with 4 attention heads.
We present more details about our model architecture in the appendix.

\noindent \textbf{Training details.} 
We take the image crop of the hand-object region according to their bounding boxes for DexYCB benchmark. Then, we modify camera intrinsic and extrinsic parameters~\cite{mehta2017monocular,yu2021pcls} accordingly and take the cropped image as the input to our model. The spatial size of input images is $256\times256$ for all our models. We perform data augmentation including rotation ($[-45^{\circ}, 45^{\circ}]$) and color jittering. 
During SDF training, we randomly sample 1000 points (500 points inside the mesh and 500 points outside the mesh) for the hand and the object, respectively. 
We train our model with a batch size of 256 for 1600 epochs on both ObMan and DexYCB using the Adam optimizer\cite{kingma2014adam} with 4 NVIDIA RTX 3090 GPUs. We use an initial learning rate of $1\times10^{-4}$ and decay it by half every 600 epochs.
It takes 22 hours for training on DexYCB and 60 hours on ObMan dataset.

\begin{table}[t]
\centering
\caption{Hand reconstruction performance with different hand kinematic features K$^h_*$ and visual feature V$_1$ on DexYCB dataset.\vspace{-.3cm}}
\tabcolsep=0.1cm
\begin{tabular}{cccccc} \toprule
 & Wrist only & All joints & ${\rm {{CD}}_{{h}}} \downarrow$ &${\rm {{FS}}_{{h}}}@1 \uparrow$ & ${\rm {{FS}}_{{h}}}@5 \uparrow$\\ \midrule
K$^h_1$ & $\times$&$\times$&0.364&0.154&0.764\\ 
K$^h_2$ &\checkmark&$\times$&0.344&0.167&0.776\\ 
K$^h_3$ & $\times$ &\checkmark&\bf{0.317}&\bf{0.171}&\bf{0.788}\\ \bottomrule
\end{tabular}
\label{tab_hand_kine}
\end{table}

\begin{table}[!t]
\centering
\tabcolsep=0.1cm
\caption{Object reconstruction performance with different object kinematic features K$^o_*$ and visual feature V$_1$ on DexYCB dataset.\vspace{-.3cm}}
\begin{tabular}{cccccc} \toprule
& Obj pose & Hand pose & ${\rm {{CD}}_{{o}}} \downarrow$ &${\rm {{FS}}_{{o}}}@5 \uparrow$ & ${\rm {{FS}}_{{o}}}@10 \uparrow$\\ \midrule
K$^o_1$ & $\times$&$\times$&2.06&0.392&0.660\\ 
K$^o_2$ & \checkmark&$\times$&1.93&0.396&0.668\\ 
K$^o_3$ & \checkmark&\checkmark&\bf{1.71}&\bf{0.418}&\bf{0.689}\\ \bottomrule
\end{tabular}
\label{tab_obj_kine}
\vspace{-.4cm}
\end{table}

\subsection{Ablation studies}
\label{subsec:ablation}

We carry out ablations on the DexYCB dataset to validate different components in our gSDF model. We evaluate different settings of hand kinematic features (K$^h_*$ in Table~\ref{tab_hand_kine}), object kinematic features (K$^o_*$ in Table~\ref{tab_obj_kine}), and visual features (V$_{*}$ in Table~\ref{tab_visual}). We use the asymmetric image backbone if not otherwise mentioned.

\begin{table*}[!t]
\centering
\setlength{\tabcolsep}{3pt}
\caption{Hand-object reconstruction performance with different visual features on DexYCB dataset. The visual features are combined with the best kinematic features K$^h_3$ (Table~\ref{tab_hand_kine}) and K$^o_3$ (Table~\ref{tab_obj_kine}) to reconstruct hand and object respectively.}
\vspace{-0.3cm}
\label{tab_visual}
\begin{tabular}{ccccccccccccc} \toprule
\multicolumn{3}{c}{} & \multicolumn{2}{c}{Transformer} & \multirow{2}{*}{${\rm {{CD}}_{{h}}} \downarrow$} & \multirow{2}{*}{${\rm {{FS}}_{{h}}}@1 \uparrow$} & \multirow{2}{*}{${\rm {{FS}}_{{h}}}@5 \uparrow$} & \multirow{2}{*}{${\rm {{CD}}_{{o}}} \downarrow$} & \multirow{2}{*}{${\rm {{FS}}_{{o}}}@5\uparrow$} & \multirow{2}{*}{${\rm {{FS}}_{{o}}}@10\uparrow$} & \multirow{2}{*}{${\rm {{E}}_{{h}}}\downarrow$} & \multirow{2}{*}{${\rm {{E}}_{{o}}}\downarrow$}\\
&Global & Local & Spatial & Temp. &  &  &  &  &  \\ \midrule
V$_1$ &\checkmark&$\times$&$\times$&$\times$&0.317&0.171&0.788&1.71&0.418&0.689&1.44&\bf{1.91}\\ 
V$_2$ &$\times$&\checkmark&$\times$&$\times$&0.310&0.172&0.795&1.71&0.426&0.694&1.44&1.98\\
V$_3$ &$\times$&\checkmark&\checkmark&$\times$&0.304&0.174&0.797&1.60&0.434&0.703&1.44&1.94\\
V$_4$ &$\times$&\checkmark&\checkmark&\checkmark&\bf{0.302}&\bf{0.177}&\bf{0.801}&\bf{1.55}&\bf{0.437}&\bf{0.709}&\bf{1.44}&1.96\\ \bottomrule
\end{tabular}
\end{table*}

\begin{table*}[t]
\centering
\caption{Hand-object reconstruction performance using different image backbone sharing strategies on DexYCB dataset. The ablation is carried out with visual features V$_1$ and kinematic features K$^h_3$ and K$^o_3$.}
\vspace{-0.3cm}
\begin{tabular}{ccccccccc} \toprule
Backbone & ${\rm {{CD}}_{{h}}} \downarrow$ &${\rm {{FS}}_{{h}}}@1 \uparrow$ & ${\rm {{FS}}_{{h}}}@5 \uparrow$&${\rm {{CD}}_{{o}}} \downarrow$ &${\rm {{FS}}_{{o}}}@5\uparrow$ & ${\rm {{FS}}_{{o}}}@10 \uparrow$&${\rm {{E}}_{{h}}}\downarrow$&${\rm {{E}}_{{o}}}\downarrow$\\ \midrule
Single&0.411&0.148&0.741&1.88&0.402&0.674&1.72&\bf{1.83}\\ 
Symmetric &0.324&0.168&0.779&1.84&0.405&0.672&1.46&1.93\\ 
Asymmetric &\bf{0.317}&\bf{0.171}&\bf{0.788}&\bf{1.71}&\bf{0.418}&\bf{0.689}&\bf{1.44}&1.91\\ \bottomrule
\vspace{-0.8cm}
\end{tabular}
\label{tab_model}
\end{table*}

\noindent\textbf{Hand kinematic feature.}
In Table~\ref{tab_hand_kine}, we evaluate the contribution of the proposed hand kinematic features for 3D hand reconstruction.
The model in K$^h_1$ does not use any pose priors to transform the 3D point.
The model in K$^h_2$ only uses the hand wrist pose to transform the 3D point as AlignSDF \cite{chen2022alignsdf}.
Our model in K$^h_3$ computes the transformations to all the hand joints, which achieves the best performance on all the evaluation metrics. 
Compared to K$^h_1$ without any pose priors, our model achieves more than 12\% and 9\% improvement on ${\rm {{CD}}_{{h}}}$ and ${\rm {{FS}}_{{h}}@1}$, respectively.
Compared to K$^h_2$ with only hand wrist, our model greatly reduces the hand Chamfer distance from 0.344 ${\rm cm}^{2}$ to 0.317 ${\rm cm}^{2}$, leading to 7.8\% relative gains.
These results demonstrate the significance of pose priors and the advantage of gSDF for 3D hand reconstruction.

\noindent\textbf{Object kinematic feature.} 
In Table~\ref{tab_obj_kine}, we validate the effectiveness of our proposed object kinematic feature. 
The model in K$^o_1$ does not contain any pose priors, while the model in K$^o_2$ alignes query points to the object center as in \cite{chen2022alignsdf}.
Our model in K$^o_3$ further employs the hand pose to produce the object kinematic feature, which significantly boosts the performance for the object reconstruction on different metrics. 
Compared to K$^o_2$, our proposed object kinematic feature achieves more than 11\% and 5.5\% improvement on ${\rm {{CD}}_{{o}}}$ and ${\rm {{FS}}_{{o}}@5}$, respectively.

\begin{figure}[t]
  \centering
  \includegraphics[width=0.75\linewidth]{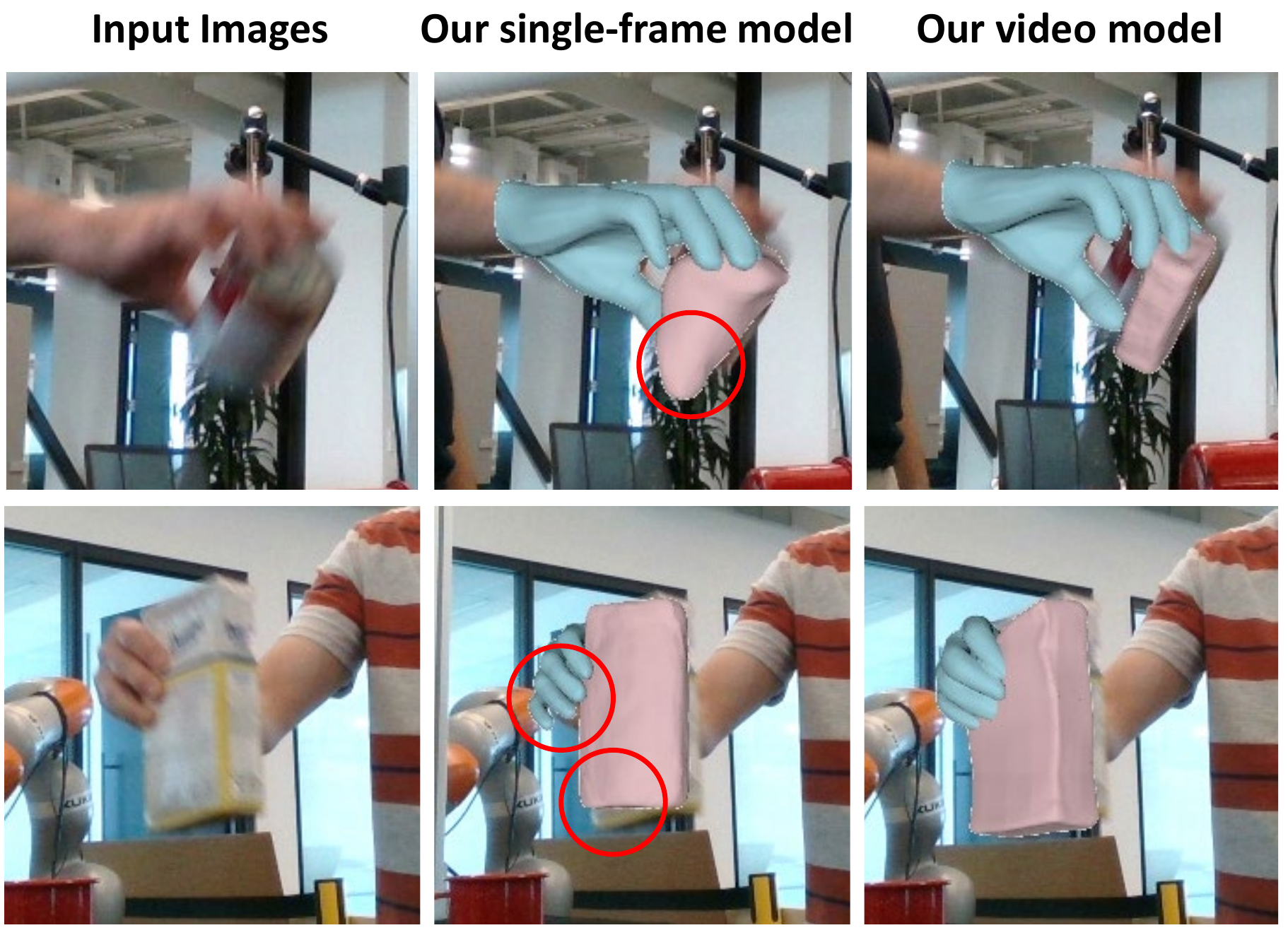}
  \vspace{-0.3cm}
  \caption{The qualitative comparsion between our single-frame model built with the transformer and our video model.}
  \vspace{-0.7cm}
  \label{fig:compare_video}
\end{figure}

\noindent \textbf{Visual features.} 
We compare different visual features for SDF prediction in Table~\ref{tab_visual}.
V$_1$ uses the global visual feature \eg the average pooling of ResNet feature map as in previous works ~\cite{karunratanakul2020grasping, chen2022alignsdf}.
Our local visual features V$_2$ derived from the geometry alignment with the query point reduces the hand Chamfer distance from 0.317 ${\rm cm}^{2}$ to 0.310 ${\rm cm}^{2}$. However, it shows less improvement on the object shape accuracy.
In V$_3$ and V$_4$, we use the transformer model to refine the feature maps.
To ablate the improvement from the transformer architecture and from the temporal information in videos, we only use transformer for each single frame in V$_3$ while use it for multiple frames in V$_4$.
We can see that the transformer architecture alone is beneficial for the reconstruction. Enhancing the visual features with temporal contexts further improves the performance in terms of all the evaluation metrics especially for the objects. In Figure~\ref{fig:compare_video}, compared with our single-frame model built with the transformer, our video model can make more robust predictions under some hard cases (\emph{e.g.,} motion blur). Although the reconstruction of the {\it can} is not accurate in the first example, our model tends to produces more regular shapes.

\noindent \textbf{Image backbone sharing strategy.} 
Results of using different strategies for image backbone sharing  are presented in Table~\ref{tab_model}. We train all the three models using the two-stage strategy described in Section~\ref{sec:method_train}. The model with one single backbone achieves the worst performance under most of the evaluation metrics. This is because the pose learning and shape learning compete with each other during training. The symmetric strategy to separate backbones for pose and SDFs performs better than the single backbone model. Our asymmetric strategy with a separate backbone for hand pose estimation and a shared backbone for object pose and SDF feature encoder achieves the best performance.
We also empirically find that learning the object pose and SDFs together improves both the pose accuracy and the shape accuracy. The possible reason is that estimating object pose also helps our model to focus more on hand-object regions and boosts the 3D reconstruction accuracy.

\begin{table*}[!t]
\centering
\small
\caption{Comparison with state-of-the-art methods on the image ObMan dataset.}
\vspace{-0.3cm}
\begin{tabular}{lcccccccc} \toprule
Methods &${\rm {CD}_{{h}}}\downarrow$ & ${\rm {FS}_{{h}}@1}\uparrow$ &${\rm {FS}_{{h}}@5}\uparrow$ & ${\rm {CD}_{{o}}}\downarrow$&${\rm {FS}_{{o}}@5}\uparrow$&${\rm {FS}_{{o}}@10}\uparrow$&${\rm {E}_{{h}}}\downarrow$&${\rm {E}_{{o}}}\downarrow$\\ \midrule
Hasson \emph{et al.}~\cite{hasson2019learning}&0.415&0.138&0.751&3.60&0.359&0.590&1.13&-\\ 
Karunratanakul \emph{et al.}~\cite{karunratanakul2020grasping} &0.261&-&-&6.80&-&-&-&-\\ 
Ye \emph{et al.}~\cite{ye2022s}&-&-&-&-&0.420&0.630&-&-\\
Chen \emph{et al.}~\cite{chen2022alignsdf}&0.136&0.302&0.913&3.38&0.404&0.636&1.27&\bf{3.29} \\ \midrule
gSDF (Ours) &\bf{0.112}&\bf{0.332}&\bf{0.935}&\bf{3.14}&\bf{0.438}&\bf{0.660}&\bf{0.93}&3.43\\ \bottomrule
\end{tabular}
\label{tab:obman}
\end{table*}
\vspace{-0.1cm}

\begin{table*}[t]
\centering
\small
\caption{Comparison with state-of-the-art methods on the video DexYCB dataset.}
\vspace{-0.3cm}
\begin{tabular}{lcccccccc} \toprule
Methods &${\rm {CD}_{{h}}}\downarrow$ & ${\rm {FS}_{{h}}@1}\uparrow$ &${\rm {FS}_{{h}}@5}\uparrow$ & ${\rm {CD}_{{o}}}\downarrow$&${\rm {FS}_{{o}}@5}\uparrow$&${\rm {FS}_{{o}}@10}\uparrow$&${\rm {E}_{{h}}}\downarrow$&${\rm {E}_{{o}}}\downarrow$\\ \midrule
Hasson \emph{et al.}~\cite{hasson2019learning}&0.537&0.115&0.647&1.94&0.383&0.642&1.67&-\\ 
Karunratanakul \emph{et al.}~\cite{karunratanakul2020grasping}&0.364&0.154&0.764&2.06&0.392&0.660&-&-\\ 
Chen \emph{et al.}~\cite{chen2022alignsdf}&0.358&0.162&0.767&1.83&0.410&0.679&1.58&\bf{1.78} \\ 
Chen \emph{et al.}~\cite{chen2022alignsdf} \footnotemark[1]$^{\dagger}$&0.344&0.167&0.776&1.81&0.413&0.687&1.57&1.93 \\ \midrule
gSDF (Ours) &\bf{0.302}&\bf{0.177}&\bf{0.801}&\bf{1.55}&\bf{0.437}&\bf{0.709}&\bf{1.44}&1.96\\ \bottomrule
\vspace{-0.8cm}
\end{tabular}
\label{tab:dexycb}
\end{table*}

\footnotetext[1]{$^{\dagger}$To make more fair comparison with Chen \etal \cite{chen2022alignsdf}, we adapt their model to the same asymmetric backbone structure as used in our method.}

\begin{figure}[!t]
  \centering
  \includegraphics[width=0.85\linewidth]{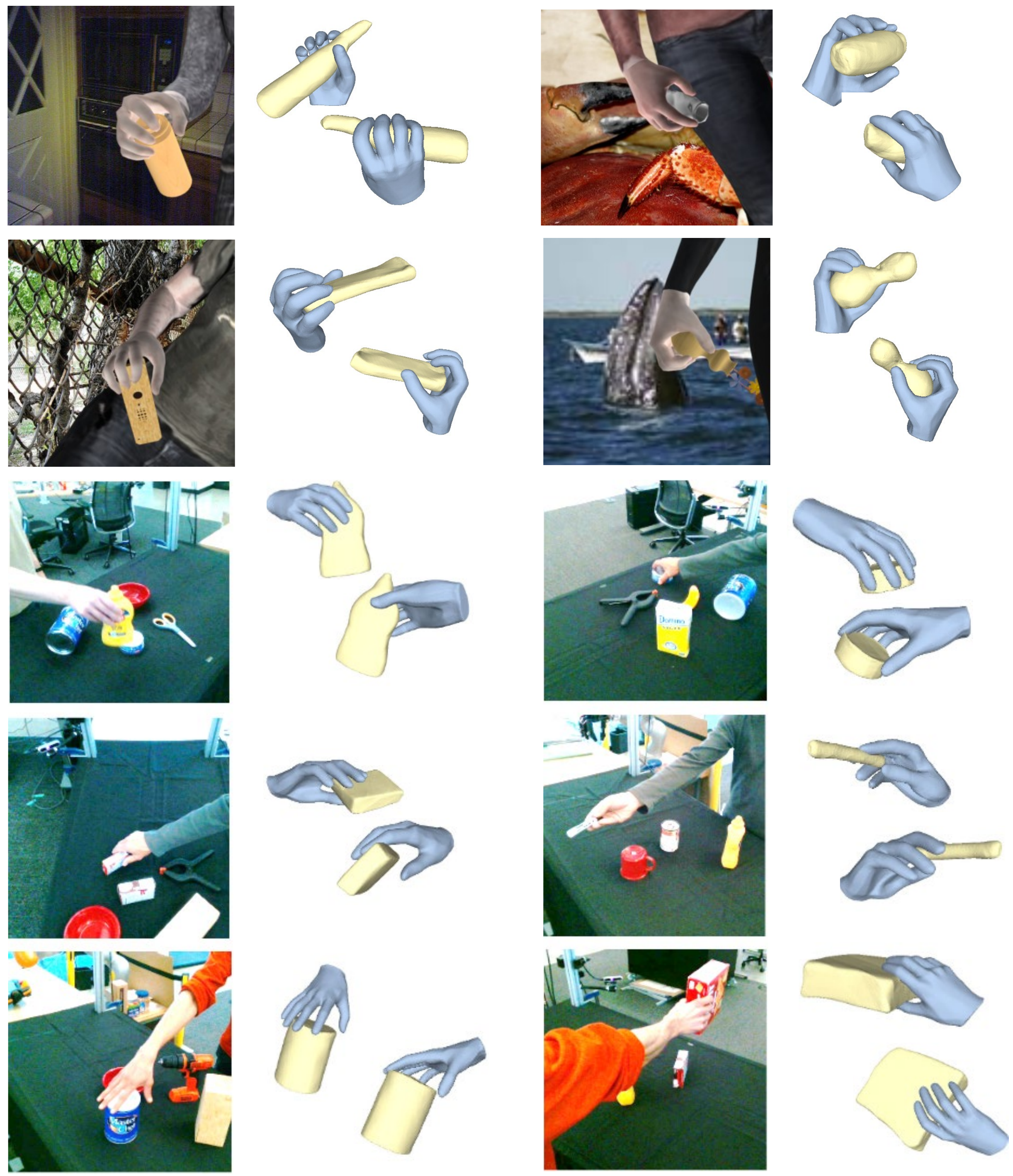}
  \vspace{-0.3cm}
  \caption{Qualitative results of our model on test images from the ObMan and DexYCB benchmarks. Our model produces convincing results for different grasping poses and diverse objects.\vspace{-.45cm}}
  \label{fig:visual}
\end{figure}

\subsection{Comparison with state of the art}
\label{subsec:sota}
We compare our gSDF model with state-of-the-art methods on ObMan and DexYCB benchmarks. In Figure~\ref{fig:visual}, we qualitatively demonstrate that our approach can produce convincing 3D hand-object reconstruction results.

\noindent \textbf{ObMan.} Table~\ref{tab:obman} shows the comparison of hand and object reconstruction results on the synthetic ObMan dataset. Since ObMan does not contain video data, we do not use the spatial-temporal transformer in this model. The proposed gSDF outperforms previous methods by a significant margin. Compared with the recent method~\cite{ye2022s} that only reconstructs hand-held objects, our joint method produces more accurate object meshes. gSDF achieves a 17.6\% improvement on ${\rm {CD}_{{h}}}$ and a 7.1\% improvement on ${\rm {CD}_{{o}}}$ over the state-of-the-art accuracy, which indicates that our model can better reconstruct both hand meshes and diverse object meshes.

\noindent \textbf{DexYCB.} Table~\ref{tab:dexycb} presents results on the DexYCB benchmark. We also show the performance of AlignSDF~\cite{chen2022alignsdf} with two backbones (\cite{chen2022alignsdf}-2BB). Our model demonstrates a large improvement over recent methods. In particular, it advances the state-of-the-art accuracy on ${\rm {CD}_{{h}}}$ and ${\rm {CD}_{{o}}}$ by 12.2\% and 14.4\%, respectively. The high accuracy of gSDF on DexYCB demonstrates that it generalizes well to real images.

%% file: 10_conclusion.tex
\vspace{-0.3cm}
\section{Conclusion}
\label{sec:conclusion}
\vspace{-0.2cm}
In this work, we propose a geometry-driven SDF (gSDF) approach for 3D hand and object reconstruction.
We explicitly model the underlying 3D geometry to guide the SDF learning.
We first estimate poses of hands and objects according to kinematic chains of pose transformations, and then derive kinematic features and local visual features using the geometry information for signed distance prediction.
Extensive experiments on ObMan and DexYCB datasets demonstrate the effectiveness of our proposed method.

{\scriptsize
\noindent \textbf{Acknowledgements.} This work was granted access to the HPC resources of IDRIS under the allocation AD011013147 made by GENCI. This work was funded in part by the French government under management of Agence Nationale de la Recherche as part of the “Investissements d’avenir” program, reference ANR19-P3IA-0001 (PRAIRIE 3IA Institute) and by Louis Vuitton ENS Chair on Artificial Intelligence. We thank Yana Hasson for helpful discussions.
}

%% file: 12_appendix.tex
\appendix
\label{sec:appendix}

In the appendix, we provide more details of our method and additional results.
We first present details of our model architecture in Section~\ref{supmat:arch}. Then in Section~\ref{supmat:ik}, we provide more details about solving hand poses from predicted 3D joints using inverse kinematics. Finally, we discuss additional experimental results in Section~\ref{supmat:exp}.

\section{Network Architecture}
\label{supmat:arch}

\begin{figure}[!ht]
  \centering  \includegraphics[width=1\linewidth]{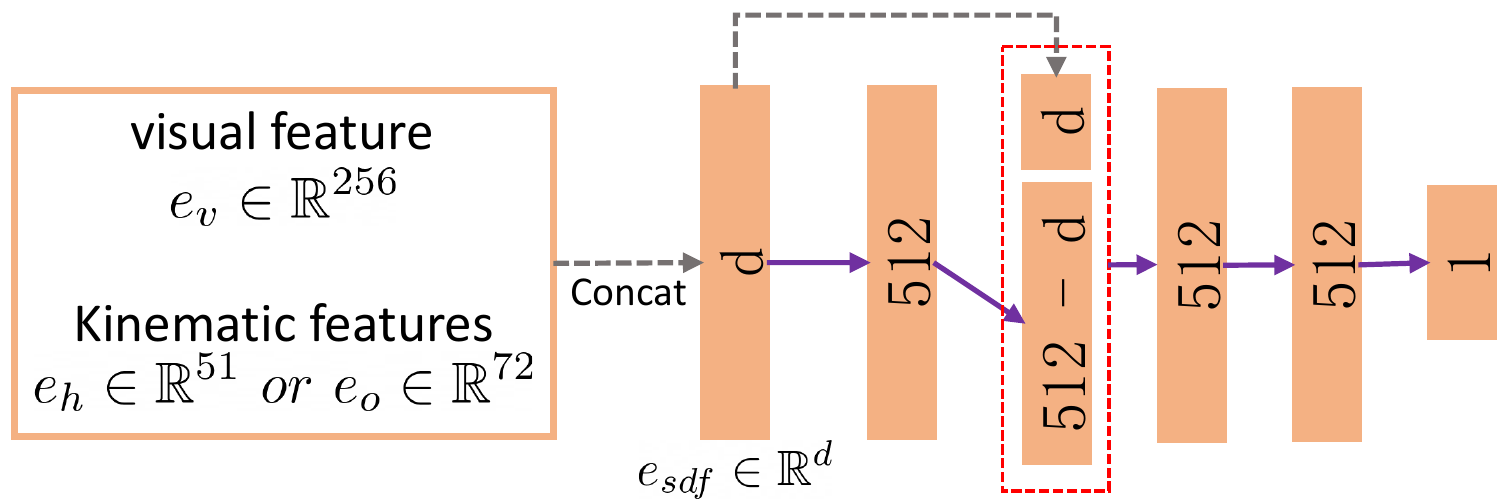}
  \caption{Network architecture used for our hand and object SDF decoders. Following~\cite{karunratanakul2020grasping,chen2022alignsdf}, we use five fully-connected layers (marked in purple) for the SDF decoder. The number in the box denotes the dimension of features.}
  \vspace{-0.35cm}
  \label{fig:supmat_arch}
\end{figure}

For our SDF decoders (see Figure~\ref{fig:method} in the original paper) we adopt the model architecture used in~\cite{karunratanakul2020grasping,chen2022alignsdf} which employ five fully-connected layers as the decoder as illustrated in Figure~\ref{fig:supmat_arch}.
Given visual feature $e_v \in \mathbb{R}^{256}$ from the input image (Section~\ref{sec:method_kinematic_embed}) and kinematic features $e_h \in \mathbb{R}^{51}$ or $e_o \in \mathbb{R}^{72}$ from the query point (Section~\ref{sec:method_visual_embed}), we concatenate them together to build a $d$-dimensional vector $e_{sdf}$ and feed it into the SDF decoder. 

\section{Hand Kinematics}
\label{supmat:ik}
\begin{figure}[t]
  \centering  \includegraphics[width=1\linewidth]{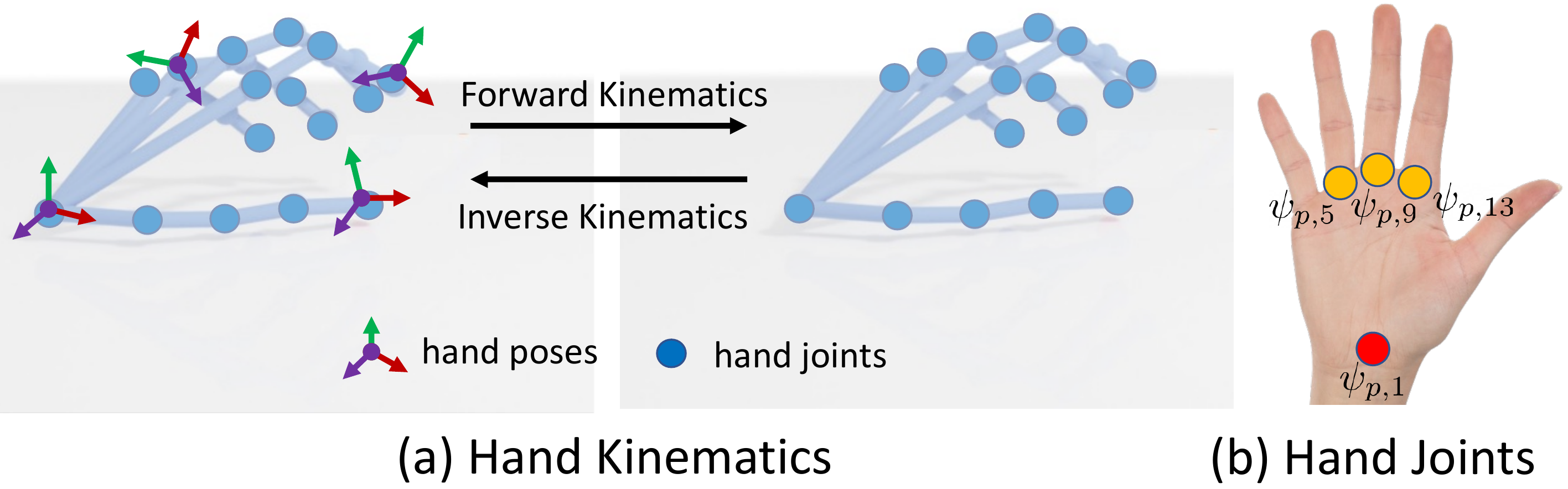}
  \caption{Illustration of hand kinematics. In Figure (a), we show functions of forward kinematics and inverse kinematics. In Figure (b), we show relevant joints (marked in yellow) that are involved in the computation of the hand wrist rotation.}
  \vspace{-0.35cm}
  \label{fig:supmat_kine}
\end{figure}

In this section, we first introduce the forward kinematics and inverse kinematics for the hand as shown in Figure~\ref{fig:supmat_kine}(a). Then we present how to use inverse kinematics to calculate hand poses from predicted 3D joints in our method.

\noindent \textbf{Forward Kinematics.}~Forward kinematics is usually defined as the process to compute posed hand joints $\psi_{p} \in \mathbb{R}^{21\times 3}$ from given hand poses (\emph{i.e.,} relative rotations $\theta \in \mathbb{R}^{16 \times 3}$ and relative translations $\phi \in \mathbb{R}^{16 \times 3}$) and template hand joints $\psi_{t} \in \mathbb{R}^{21\times 3}$. The $k_{th}$ joint in $\psi_{p}$ can be computed as:
\begin{equation}
\begin{split}
&\psi_{p, k} = R_{k} \cdot \phi_{k} + \psi_{p,pa(k)},\\
&R_{k} = R_{pa(k)} \cdot {\rm {exp}}(\theta_{k}),
\end{split}
\label{ik_initial}
\end{equation} where $R_{k}$ denotes the global rotation matrix for the $k_{th}$ joint and $pa(\cdot)$ returns the parent index of the $k_{th}$ joint. ${\rm exp}(\cdot)$ denotes \emph{Rodrigues formula} to convert $\theta_{k}$ into the form of the rotation matrix. We follow the inverse order of the kinematic chain to derive the global rotation for the $k_{th}$ joint. For simplicity, we assume that all hands share the same template and set the relative translation as  $\phi_{k} = \psi_{t, k} - \psi_{t, pa(k)}$, which simplifies the computation of Equation~\ref{ik_initial} to: 
\begin{equation}
\begin{split}
&\psi_{p, k} = R_{k} \cdot (\psi_{t, k} - \psi_{t, pa(k)}) + \psi_{p,pa(k)},\\
&R_{k} = R_{pa(k)} \cdot {\rm {exp}}(\theta_{k}). 
\end{split}
\label{foward_kine}
\end{equation}

\noindent \textbf{Inverse Kinematics.} Given posed hand joints $\psi_{p}$ and template hand joints $\psi_{t}$, inverse kinematics solves relative hand poses ($\theta$, $\phi$) that defines the transformations from $\psi_{t}$ to $\psi_{p}$. As we do in forward kinematics, we also omit $\phi$ in the computation of inverse kinematics and only solves relative hand rotations $\theta$. We first derive the hand wrist rotation matrix $R_{1} \in \mathbb{R}^{3\times3}$ from the orientation of three connected joints as shown in Figure~\ref{fig:supmat_kine}(b) and formulate it as an optimization problem:
\begin{equation}
   R_1 = \arg\min_{R \in \mathbb{SO}^3} \sum_{i\in \{5, 9, 13\}} \Big \|\psi_{p,i} - R \cdot \psi_{t,i} \Big \|_2^2,
\end{equation} where we can apply Singular Value Decomposition (SVD) as in \cite{li2021hybrik} to solve this problem. Then, we follow the hand kinematic chain and solve the 3D rotation recursively for each joint. To this end, we rewrite Equation~\ref{foward_kine} defined in forward kinematics:
\begin{equation}
\begin{split}
R_{pa(k)}^{-1}(\psi_{p, k} - \psi_{p,pa(k)})= {\rm {exp}}(\theta_{k}) (\psi_{t, k} - \psi_{t, pa(k)}).\\
\end{split}
\end{equation}
Then, we could derive the norm and orientation of $\theta_{k}$ by computing the dot product and cross product between the vector $R_{pa(k)}^{-1}(\psi_{p, k} - \psi_{p,pa(k)})$ and the vector $\psi_{t, k} - \psi_{t, pa(k)}$, respectively.

\begin{table}[t]
\centering
\caption{Comparison with state-of-the-art methods on ObMan.}
\begin{tabular}{lccc} \toprule
Method &${\rm {C}_{{r}}}$ & ${\rm {P}_{{d}}}$ &${\rm {I}_{{v}}}$\\ \midrule
Hasson \emph{et al.}~\cite{hasson2019learning}&94.8\%&1.20&6.25\\
Karunratanakul \emph{et al.}~\cite{karunratanakul2020grasping}&69.6\%&0.23&0.20\\ 
Chen \emph{et al.}~\cite{chen2022alignsdf}&95.5\%&0.66&2.81\\ \midrule
gSDF (Ours)&89.8\%&0.42&1.17\\ \bottomrule
\end{tabular}
\label{tab:supobman}
\end{table}

\begin{table}[!t]
\centering
\caption{Comparison with state-of-the-art methods on DexYCB.}
\begin{tabular}{lccc} \toprule
Method &${\rm {C}_{{r}}}$ & ${\rm {P}_{{d}}}$ &${\rm {I}_{{v}}}$\\ \midrule
Hasson \emph{et al.}~\cite{hasson2019learning}&95.7\%&1.15&9.64\\
Karunratanakul \emph{et al.}~\cite{karunratanakul2020grasping}&96.0\%&0.92&6.62\\ 
Chen \emph{et al.}~\cite{chen2022alignsdf}&96.6\%&1.08&8.40\\ \midrule
gSDF (Ours)&95.4\%&0.94&6.55\\ \bottomrule
\end{tabular}
\label{tab:supdexycb}
\end{table}


\begin{table}
\centering
\tabcolsep=0.1cm
\caption{Object reconstruction performance with different object kinematic features on DexYCB dataset. $^*$ denotes our re-implementation of the method proposed in Ye~\emph{et al.}~\cite{ye2022s}.}
\label{tab:supcomp}
\begin{tabular}{lccccc} \toprule
 & Model & Obj. Pose & ${\rm {{CD}}_{{o}}} \downarrow$ &${\rm {{FS}}_{{o}}}@5 \uparrow$ & ${\rm {{FS}}_{{o}}}@10 \uparrow$\\ \midrule
R1 & Ye~\emph{et al.}~\cite{ye2022s} & $\times$ &-&0.420&0.630  \\
R2 & Ye~\emph{et al.}$^*$ & $\times$ &2.09&0.404&0.663  \\ \midrule
R3 & \multirow{2}{*}{\begin{tabular}[c]{@{}l@{}}gSDF\\ (Ours)\end{tabular}}  & $\times$ &1.78&0.411&0.676  \\
R4 &  & \checkmark &\bf{1.71}&\bf{0.418}&\bf{0.689} \\ \bottomrule
\end{tabular}
\end{table}

\begin{table}[!ht]
\centering
\caption{Comparing computational requirements of different models when reconstructing hand and object meshes of resolution $128\times128\times128$ from an image on an NVIDIA 1080Ti GPU.}
\setlength{\tabcolsep}{4.5pt}
\renewcommand\arraystretch{0.9}
\begin{tabular}{cccc} \toprule
Method & Input & GPU Memory & Latency\\ \midrule 
\cite{karunratanakul2020grasping}&Image&2357Mb&2.87s\\ 
\cite{chen2022alignsdf}&Image&2847Mb&3.17s\\ 
Ours&Image&3425Mb&3.23s\\
Ours&Video&3764Mb&4.14s\\
\bottomrule
\end{tabular}
\label{tab_resource}
\end{table}

\begin{table}[!th]
\centering
\footnotesize
\caption{Comparision of our method with AlignSDF~\cite{chen2022alignsdf} on DexYCB while using different numbers of backbones (BB).}
\vspace{-0.3cm}
\setlength{\tabcolsep}{1.3pt}
\renewcommand\arraystretch{1.0}
\begin{tabular}{ccccccc} \toprule
Model & ${\rm {{CD}}_{{h}}} \downarrow$ &${\rm {{FS}}_{{h}}}@1 \uparrow$ & ${\rm {{FS}}_{{h}}}@5 \uparrow$&${\rm {{CD}}_{{o}}} \downarrow$ &${\rm {{FS}}_{{o}}}@5\uparrow$ & ${\rm {{FS}}_{{o}}}@10 \uparrow$\\ \midrule 
\cite{chen2022alignsdf}-1BB&0.358&0.162&0.767&\textbf{1.83}&0.410&0.679\\ 
Ours-1BB&\textbf{0.329}&\textbf{0.166}&\textbf{0.787}&1.88&\textbf{0.420}&\textbf{0.689}\\ \midrule
\cite{chen2022alignsdf}-2BB&0.344&0.167&0.776&1.81&0.413&0.687\\ 
Ours-2BB&\textbf{0.310}&\textbf{0.172}&\textbf{0.795}&\textbf{1.71}&\textbf{0.426}&\textbf{0.694}\\ \midrule
Ours-3BB&0.326&0.168&0.784&1.82&0.414&0.679 \\
\bottomrule
\end{tabular}
\vspace{-0.6cm}
\label{tab_new_model}
\end{table}

\section{Experimental Results}
\label{supmat:exp}
\subsection{Evaluations using additional metrics}
To provide a more comprehensive view about our 3D reconstruction performance, we also report Contact Ratio (${\rm {C}_{r}}$), Penetration Depth (${\rm {P}_{d}}$) (${\rm cm}$) and Intersection Volume (${\rm {I}_{v}}$) (${\rm cm}^{3}$) for our models. We follow the same process as previous works~\cite{karunratanakul2020grasping,chen2022alignsdf} to compute these metrics. As shown in Table~\ref{tab:supobman} and  Table~\ref{tab:supdexycb}, we can observe that our approach can generate results with relatively low Penetration Depth (${\rm {P}_{d}}$) and Intersection Volume (${\rm {I}_{v}}$) on both the ObMan and DexYCB benchmarks, which suggests that our model can produce physically plausible 3D reconstruction of hand and object meshes. Table~\ref{tab_resource} compares the speed and memory of different models.
Our image model only slightly increases compute compared to \cite{karunratanakul2020grasping,chen2022alignsdf}.

\begin{figure*}[t]
  \centering  \includegraphics[width=0.86\linewidth]{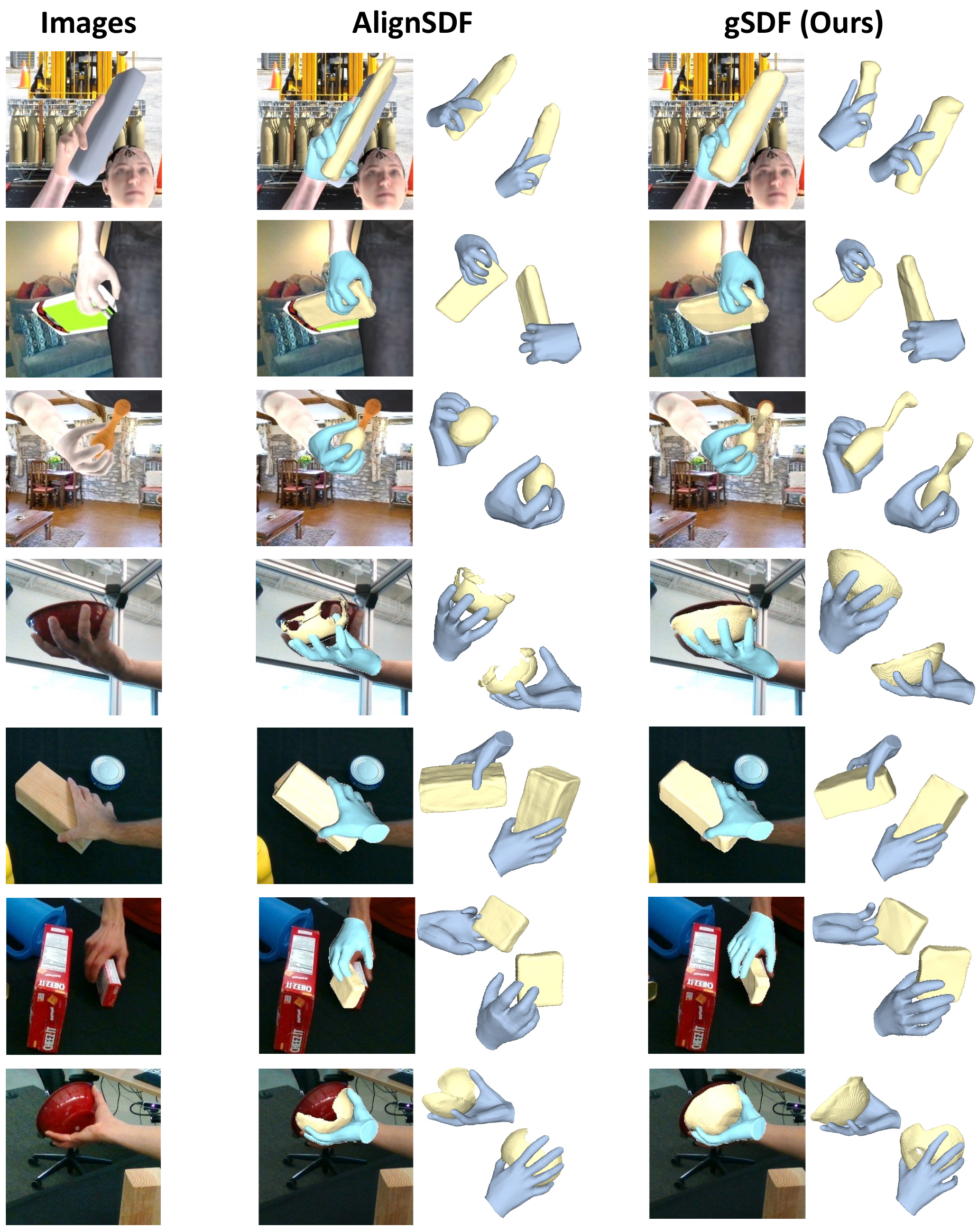}
  \caption{Qualitative comparison between AlignSDF~\cite{chen2022alignsdf} and our gSDF. Our approach can produce more realistic hand and object reconstruction results.}
  \label{fig:supmat_cmp}
\end{figure*}

\begin{figure}[t]
  \centering  \includegraphics[width=0.9\linewidth]{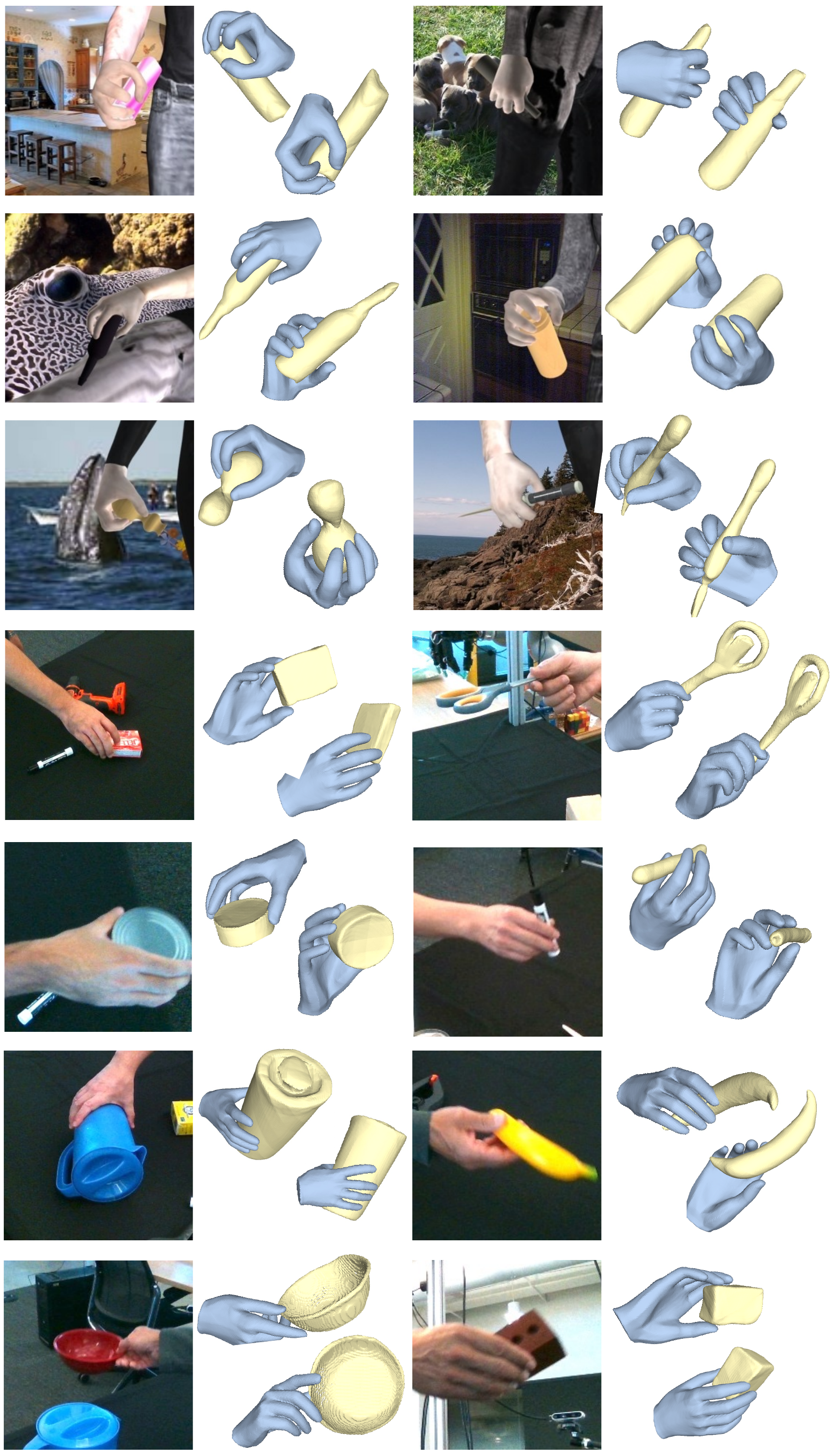}
  \caption{Qualitative results of our model on test images from the ObMan and DexYCB benchmarks. Our approach can produce convincing 3D reconstruction results for different hand grasping poses and challenging objects.}
  \label{fig:supmat_demo}
\end{figure}

\begin{figure}[!tp]
  \centering
  \includegraphics[width=0.48\textwidth]{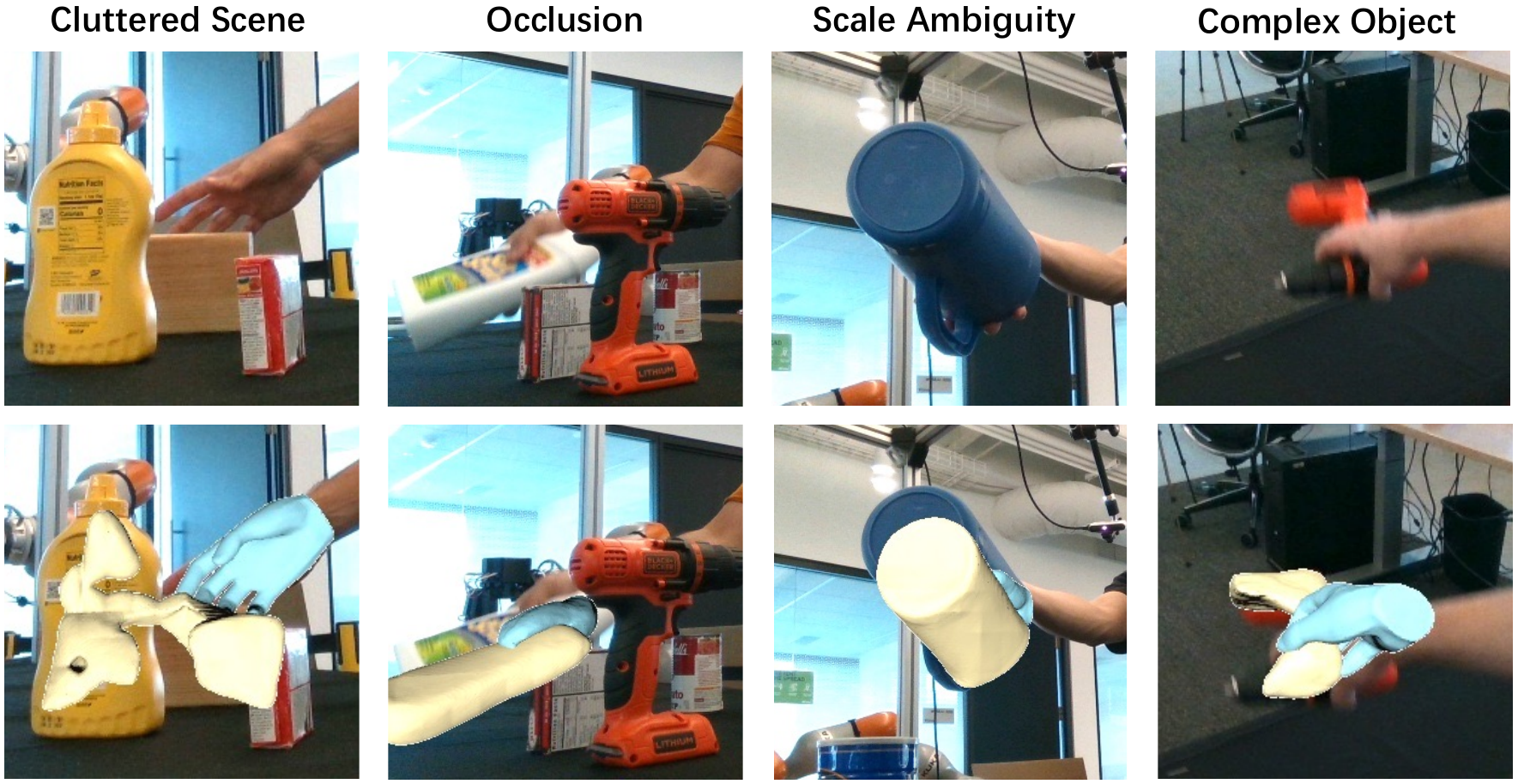}
  \caption{Failure cases analysis of our method on the DexYCB benchmark.}
  \label{failure_analysis}
\end{figure}

\subsection{Comparison with Ye \etal~\cite{ye2022s}}
As Ye \etal~\cite{ye2022s} is a close work related to ours, we provide more ablation results for comparison with Ye \etal~\cite{ye2022s} in Table~\ref{tab:supcomp}.
The main differences between Ye \etal~\cite{ye2022s} and our work are three-fold. Firstly, they focus on 3D hand-held object reconstruction instead of joint hand-object reconstruction. Secondly, they only consider the hand poses for object reconstruction without object poses, and the hand poses are predicted from an off-the-shelf model. Finally, a larger SDF decoder is used in their work while we follow \cite{karunratanakul2020grasping,chen2022alignsdf} and use a smaller decoder architecture.
Therefore, in Table~\ref{tab:supcomp}, we only compare the object reconstruction performance.
We also re-implement Ye \etal~\cite{ye2022s} (R2 in Table~\ref{tab:supcomp}) using the same SDF decoder and the same predicted hand poses as ours for a fair comparison.
The model in R3 indicates that the joint optimization of hand-object reconstruction is beneficial compared to the model in R2.
Our model in R4 uses both hand poses and object poses to produce object kinematic features and achieves the best performance on all the metrics for 3D object reconstruction. 

\subsection{Ablations on the number of backbones}
Table~\ref{tab_new_model} reports additional results showing improvements of our method over~\cite{chen2022alignsdf} while using the same number of backbones. We note that all models in this table are trained with the local visual features V$_2$ defined in Table~\ref{tab_visual}. We observe that gSDF can still outperform AlignSDF~\cite{chen2022alignsdf} under a single backbone setting. For a better comparison, we also extend AlignSDF to two backbones and train it with the two-stage strategy. 2BB results in Table~\ref{tab_new_model} show that our method outperforms~\cite{chen2022alignsdf} even when both methods use two backbones. We further conduct an experiment with three backbones, where we use three separate backbones for hand and object pose estimation and SDF learning. We observe that 3BB consumes more resources without improving performance. This shows that object pose estimation and SDF learning benefit from a shared backbone in our 2BB asymmetric architecture.

\subsection{Qualitative results}
In this section, we include more qualitative examples in Figure~\ref{fig:supmat_demo} to show that our approach can reconstruct high-quality hand meshes and object meshes for some challenging cases. We also qualitatively compare our method with a most recent work AlignSDF~\cite{chen2022alignsdf} on both the ObMan and DexYCB benchmarks. As shown in Figure~\ref{fig:supmat_cmp}, we can observe that our method produces more realistic reconstruction results. Even for some objects with thin structures (\emph{e.g.}, bowl), our method can still faithfully recover their 3D surfaces. 

\subsection{Failure cases analysis} 
In this section, we analyze some typical patterns for our method on the DexYCB benchmark. As shown in Figure~\ref{failure_analysis}, our method sometimes makes unreliable predictions in cluttered scenes. Our method uses a hand-relative coordinate system. Hence, the reconstruction of both hands and objects may fail for scenes with heavily occluded hands. Since our method takes monocular RGB frames as the input, reconstructed objects, especially for big objects, might have incorrect scales. For some objects with complex geometric topology, it is still difficult to produce accurate 3D reconstructions under strong motion blur.